# A Knowledge-based Approach for Answering Complex Questions in Persian

Romina Etezadi[a], Mehrnoush Shamsfard[a,*]

[a]*Shahid Beheshti University, Tehran, Iran*

**Abstract**

Research on open-domain question answering (QA) has a long tradition. A challenge in this domain is answering complex questions (CQA) that require complex inference methods and large amounts of knowledge. In low resource languages, such as Persian, there are not many datasets for open-domain complex questions and also the language processing toolkits are not very accurate. In this paper, we propose a knowledge-based approach for answering Persian complex questions using *Farsbase*; the Persian knowledge graph, exploiting *PeCoQ*; the newly created complex Persian question dataset. In this work, we handle multi-constraint and multi-hop questions by building their set of possible corresponding logical forms. Then Multilingual-BERT is used to select the logical form that best describes the input complex question syntactically and semantically. The answer to the question is built from the answer to the logical form, extracted from the knowledge graph. Experiments show that our approach outperforms other approaches in Persian CQA.

## 1. Introduction

Having a system capable of answering real-world natural language questions is still an open problem that has attracted AI researchers for many years. Knowledge-based question answering (KBQA) is the task of answering questions using structured resources such as Freebase (Bollacker et al., 2008), DBpedia (Auer et al., 2007), and Wikidata (Vrandei and Krtzsch, 2014). There are many studies for answering simple questions that can be answered by querying a single triple in the KB (Yih et al., 2014; Bordes et al., 2015; Dong et al., 2015; Hao et al., 2017). However, complex questions have multiple semantic constraints that we need to apply to find the correct answer(s).

There are two main types of complexities, a complex question may have: multi-hops and multi-constraints. Multi-hop questions are those that need to move through a sequence of relations in the knowledge graph to find the answer (e.g. *writer_of* and *death_cause* in the first row of Table 1). Multi-constraint questions should satisfy more than one constraint to find the answer. The constraints can be classified into four classes: multi-entity, type, temporal (implicit and explicit), and operational (aggregation, superlative, comparative). Examples for each type are shown in Table 1.

Most recent approaches try to convert the given complex question into its corresponding logical form using semantic parsing methods. However, there are several challenges in the task of semantic parsing. Traditional semantic parsing methods require ontology matching as the logical form predicates differ from ones in the KB (Kwiatkowski et al., 2013). Recent approaches solve ontology matching challenges by formulating the semantic parsing to a search problem that leverages the knowledge base more tightly (Bao et al., 2014; Yih et al., 2015; Bao et al., 2016; Luo et al., 2018; Hu et al., 2018; Lan and Jiang, 2020). There are two challenges in the task of search problems: (1) Controlling and restricting the search space, and (2) Finding the correct candidate among those produced in the search space. For the first challenge, different techniques such as defining constraints and using reinforcement learning methods are used (Tugwell, 1995; Lan and Jiang, 2020). Recent papers proposed linear scoring functions that get important features of candidates and assign a score to each (Yih et al., 2015; Bao et al., 2016; Luo et al., 2018; Hu et al., 2018; Lan and Jiang, 2020).

In this paper, we propose a method that maps the question to its corresponding logical form (SPARQL),

*Corresponding Author

*Email addresses:* ro.etezadi@mail.sbu.ac.ir (Romina Etezadi), m-shams@sbu.ac.ir (Mehrnoush Shamsfard)

Table 1: Complex question types with example.

| Complexity Type | Example |
|---|---|
| Multi-hop | What was the **death cause** of the **writer of** the Cranberries? |
| Multi-entity | What movies have **Ben Affleck** and **Matt Damon** appeared in together? |
| Type | What **rivers** run through Rome? |
| Temporal-explicit | Who was the United States president in **1990**? |
| Temporal-implicit | How old was Tom Cruise **when Lewis Allen died**? |
| Operational-aggregation | **How many** books did Alexandre Dumas write? |
| Operational-superlative | What is **the second** tallest mountain in the world? |
| Operational-comparative | Which mountains are **heigher** than Lhotse? |

considering both types of complexities. Our method focuses on dealing with complexities in Persian and creates a set of possible logical forms for the given complex question. For choosing the best logical form, we adopt Multilingual-BERT (Devlin et al., 2018). A question may have more than one operational constraint. The order of applying operational constraints affects the accuracy of the answers. Therefore, we utilize a dependency parser to check which one should be applied first. Information required for each constraint is extracted from Farsbase (Sajadi et al., 2018) and is combined to create a set of possible logical forms. Experiments show that our method achieves the state-of-the-art on PeCoQ (Etezadi and Shamsfard, 2020) dataset. Our contribution is twofold: (1) A novel approach for Persian complex question answering based on dependency parsing and logical form generation. (2) Using the BERT model instead of linear scoring functions as the ranker for choosing the best logical form. The remainder of this paper is organized as follows: In Section 2 we address the related work. Section 3 talks about the resources that we are using for Persian QA. Section 4 describes our proposed approach. The results and experiments are presented in Section 5, and finally, Section 6 concludes the paper.

## 2. Related Work

One of the important aspects of QA systems is the type of resource they use to find answers. Knowledge bases and raw texts are widely used for extracting information. There are also QA systems that utilize both types of resources (Bordes et al., 2014; Dong et al., 2015; Xu et al., 2016; Sun et al., 2019). Knowledge base question answering (KBQA) has attracted more attention in recent years. The most popular methods for KBQA can be divided into three classes: information retrieval based, embedding based and semantic parsing. Information retrieval-based approaches try to find the correct answers to the given question by exploring the knowledge graph and raw texts directly. The main challenge in this approach is finding the correct answers among the candidate answers retrieved from KG. There are various methods presented to deal with this challenge (Bordes et al., 2014; Dong et al., 2015; Xu et al., 2016; Sun et al., 2019). Embedding-based approaches learn low-dimensional vectors for both words and KGs, which allow inferring with those vectors for finding the answers (Chen et al., 2019; Saxena et al., 2020). Embedding-based approaches mostly concentrate on handling multi-hop complexity, especially where all the information is not in the knowledge graph (incompleteness problem). For example, Saxena et al. (2020) proposed a system that can find answers from half masked KG based on question and knowledge graph embeddings. Meanwhile, semantic parsing approaches try to convert the natural language question into its corresponding logical forms such as $\lambda$-DCS[1] (Liang, 2013; Berant et al., 2013; Berant and Liang, 2014), query graph (Yih et al., 2015; Bao et al., 2016; Luo et al., 2018; Hu et al., 2018; Lan and Jiang, 2020) or SPARQL (Unger et al., 2012; Yahya et al., 2012; Bast and Haussmann, 2015; Zhang et al., 2019). Building query graphs or $\lambda$-DCS need an additional step to convert them into executable queries like SPARQL. On the other hand, some systems focus on producing the corresponding SPARQL directly. For example, In Zhang et al. (2019), an end-to-end semantic parser was introduced based on a deep model that tries to decompose a given complex question and produce its SPARQL.

Most studies have relied on English QA systems. Only a few works in the literature demonstrate methods to build QA systems to answer Persian questions. Persian QA systems are mostly created on closed-domain questions (Boreshban and Mirroshandel, 2017; Veisi and Shandi, 2020) or only considering simple questions (Veyseh, 2016). To the best of our knowledge, there are no previous studies on answering open-domain Persian complex questions using KB. As we are working in a low-resource setting, there are not many related large datasets.

[1] Lambda dependency-based compositional semantics

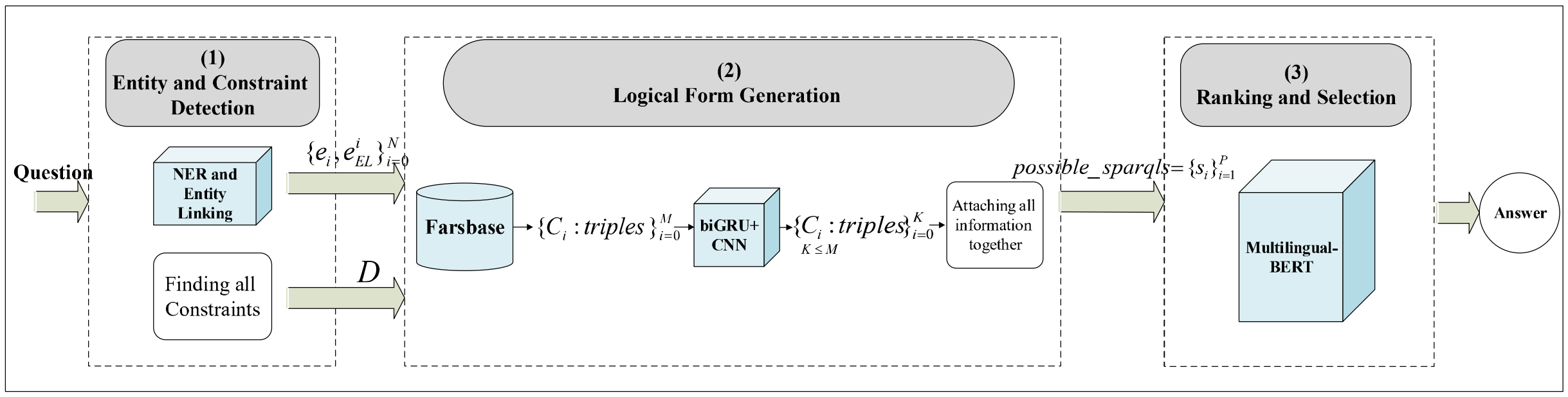

Figure 1: The main workflow of our methods can be divided into three parts. In the first part, constraints ($D$) in the question and entities ($\{e_i\}_{i=0}^{N}$) with their corresponding entity links ($\{e_{EL}^{i}\}_{i=0}^{N}$) to the knowledge graph are extracted. In the second part, we extract information from KG for each constraint $C$ ($\{C_i : triples\}_{i=0}^{M}$) and build the possible logical forms ($\{s_i\}_{i=0}^{P}$) based on the information extracted from the previous step. Finally, we rank each logical form and select the best one to execute against KG.

In this paper, we present a semantic parsing-based system that converts complex questions into SPARQL. The flow of our method is similar to the works (Bao et al., 2016; Luo et al., 2018). However, there are differences between our method and previous ones: (1) we deal with the Persian language, (2) we cover more constraints such as comparative constraint, (3) we decompose operational phrases for questions with multiple operational constraints, (4) we adopt Multilingual-BERT instead of using linear scoring function (Bao et al., 2016; Luo et al., 2018) for choosing the best one among candidate SPARQLs. Since there are no previous works on Persian complex question answering, we use Zhang et al. (2019) system to see how well our method works in generating SPARQL compare to theirs.

## 3. Resources

Various types of structured resources are used for QA tasks, such as curated KBs (e.g. knowledge graphs) (Hu et al., 2018; Lan and Jiang, 2020), open KBs (Khot et al., 2017; Kwon et al., 2018) or tables (Herzig et al., 2020). In this paper, we use Farsbase, The Persian knowledge graph, to answer the questions. In addition, the Persian complex-question dataset, PeCoQ, is used for training deep models. We describe these two resources in sections 3.1 and 3.2.

### *3.1. Farsbase*

Farsbase (Sajadi et al., 2018) is a Persian knowledge graph whose data is extracted from the Persian Wikipedia. Data in Farsbase are in the form of subject-relation-object triples ($s$, $r$, $o$), where $s$ and $o$ are entities, representing the nodes, and $r$ is a relation that connects $s$ to $o$, representing directed edges in graph. In contrary to Freebase, Farsbase does not have any CVT[2] nodes.

### *3.2. PeCoQ*

We use the complex question dataset, PeCoQ (Etezadi and Shamsfard, 2020), to train models and evaluate our method. PeCoQ has about 10,000 complex questions (8000 train, 1000 dev, and 1000 test). Complex questions in PeCoQ have different hops (up to two) and different constraints: Temporal (implicit and explicit), superlative, comparative, aggregation, multi-entity. PeCoQ is created automatically based on the information in Farsbase. Then it is revised manually and for each question, two paraphrases are added by linguists.

## 4. The Proposed Approach

There are three main steps in our approach to answer a complex question: 1) Entity and Constraint Detection. 2) Logical Form Generation. 3) Ranking and Selection. The diagram of our workflow is shown in Figure 1.

[2]Stands for component value type. CVT node is not a real-world entity. It is only used to collect multiple fields of an event.

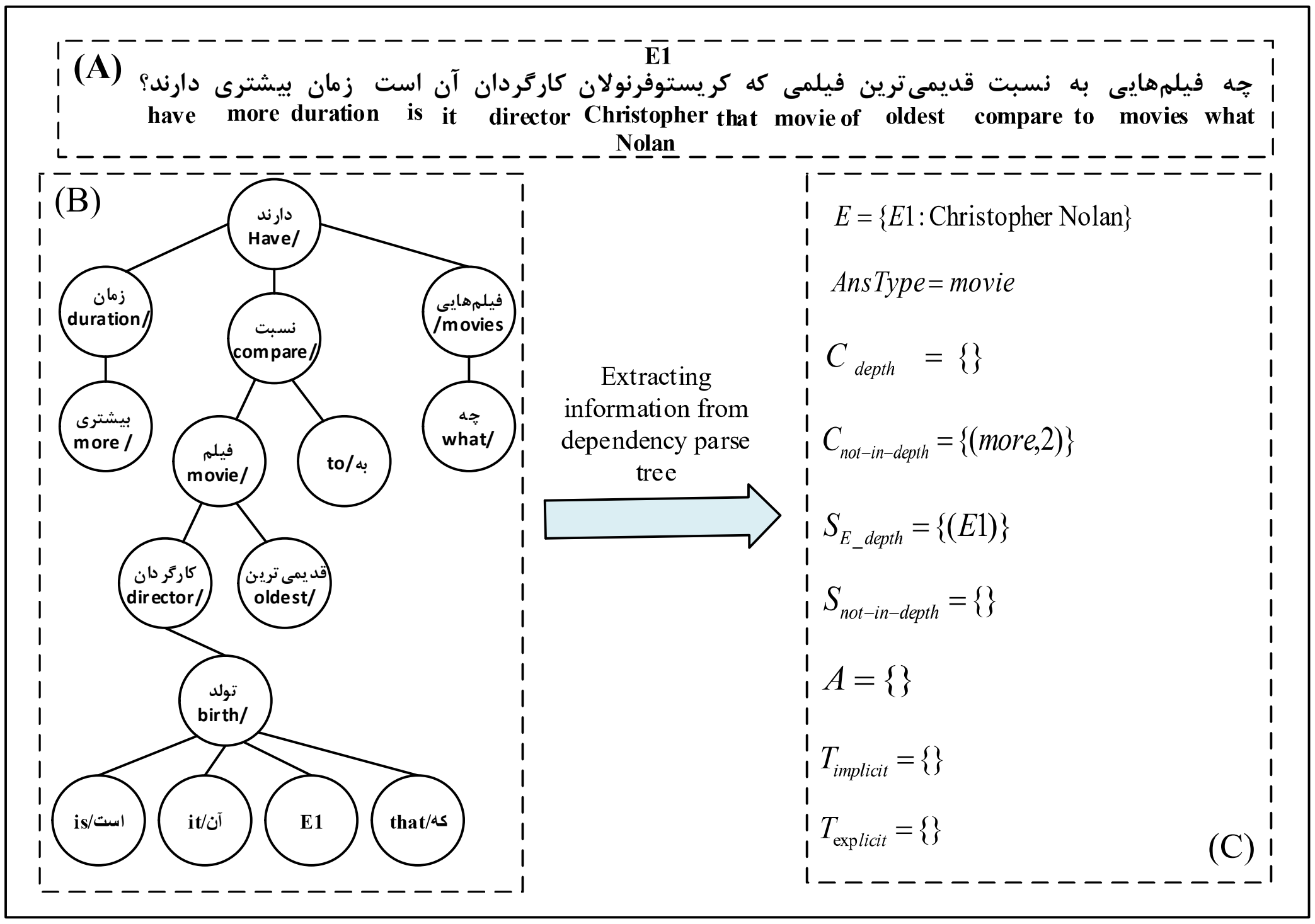

Figure 2: Example of extracting information (detecting entity and constraints) from the question. In (A), the input is question *"Which movies are longer than oldest movie directed by Christopher Nolan?"*. English translation of each Persian word is written below it. The entity in the question is found and is replaced by E1. In (B), the corresponding dependency tree (created by stanza) is shown. Information required for generating logical form is shown in (C).

### *4.1. Entity and Constraint Detection*

A complex question may have the following constraints: multi-entity, type, temporal (implicit and explicit), and operational (aggregation, superlative, and comparative). In the first step, the entities should be extracted from the question and added to the set $E$. for this purpose a named entity extraction module is developed by fine-tuning ParsBERT[3] (Farahani et al., 2020) over PeCoQ. The module tags the question by entity IOB tags (*B-E* for begin of entity, *I-E* for inside entity and *O* for outside). Experimental results show that our proposed NER method outperforms the existing Persian NER tools for this task. The evaluation of our NER tool is addressed in Section 5. Then we develop an entity linking module using the convolutional neural network (CNN) model proposed by Yih et al. (2014). Using this module, for each entity in the question, we find the corresponding entity in Farsbase with a confidence score. We then replace all the entities in the question $Q$ with dummy token $E_i$ (*ith* entity) to create $Q'$. To extract constraints from the question, we define some rules on the dependency tree of $Q'$ using *stanza* (Qi et al., 2020). Extracted constraints are added to a list to be applied later in logical form generation in Section 4.2. It is worth mentioning that constraint detection rules we discuss in this paper are for the Persian language. These rules are listed as follows.

*Explicit Temporal Constraint.* To find explicit temporal, we use the NER tool to extract explicit temporal time tokens (e.g. 2001 in *"Who was the president of the US in 2001"*).

*Implicit Temporal Constraint.* For this constraint, we assume that each time phrase should have an entity. Hence, first, we detect the subtree of the time phrase based on time keywords and then extract the entity that appears in the time phrase subtree (e.g. the entity *Lincoln* and time keyword *when* in the time phrase *"when*

[3]https://github.com/hooshvare/parsbert

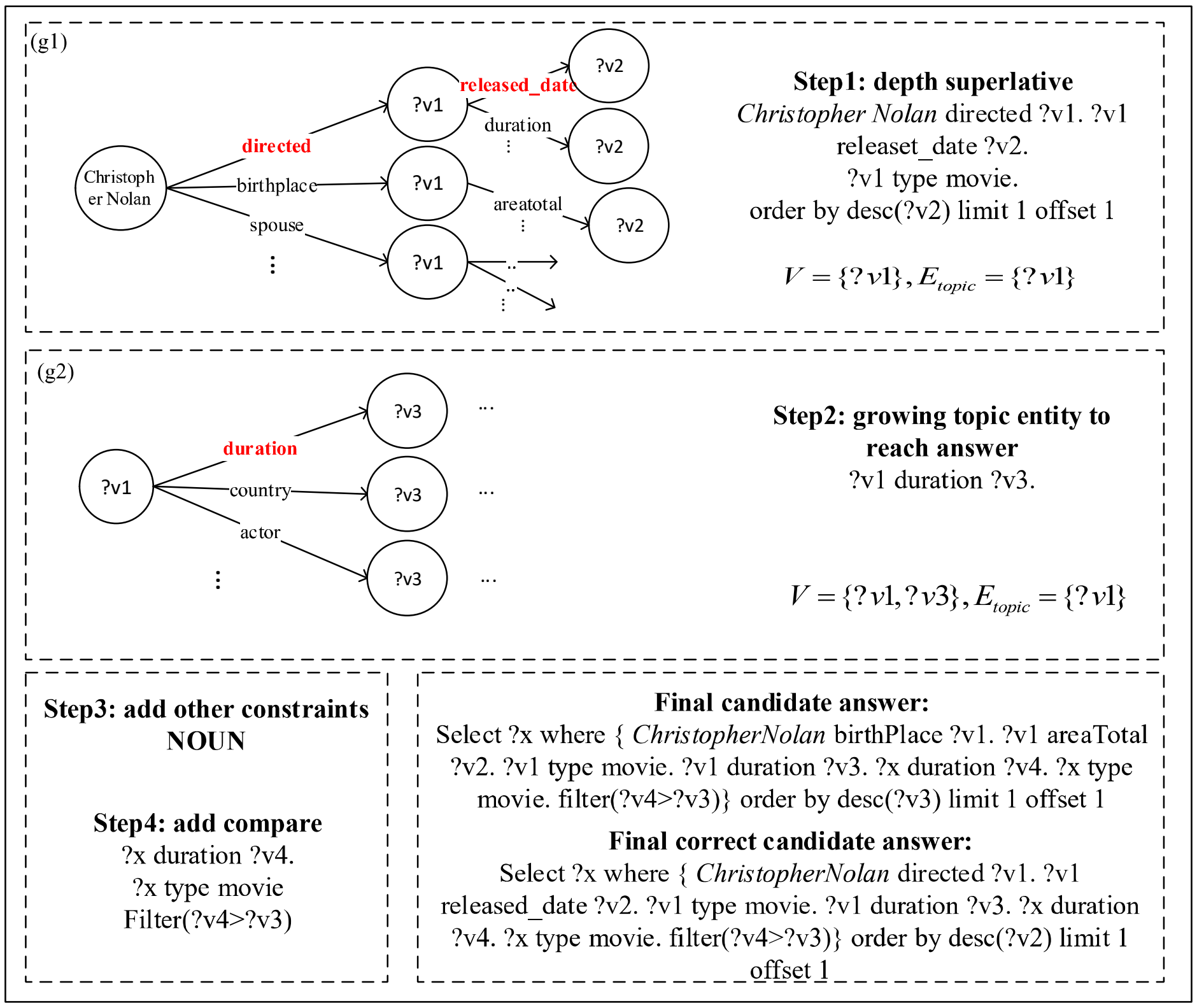

Figure 3: Steps for generating candidate logical forms.

*Lincoln was born"*).

*Type Constraint.* To find the answers' type, we use either the head token of the question word in the dependency tree (e.g. river in *which river*) or the question word itself (e.g., *who* indicates a person).

*Aggreagtion Constraint.* An aggregation constraint is used when there is an aggregate word in the question such as "how many", "number of", etc.

*Superlative Constraint.* For superlative constraints, the apply order is important in finding the correct answer. For example, in the question, *"Which is the oldest movie among movies that are longer than Tenet?"* the comparative constraint (longer) should be applied first to extract a list of movies and then the superlative constraint (oldest) should be applied on the results, while in *"Which movies are longer than the oldest movie directed by Christopher Nolan?"*, the apply order is reverse (superlative before comparative). We find each superlative phrase in the question based on the position of the superlative keyword in the dependency tree. Therefore, for each superlative keyword ($S_i$) found in the question, we find its depth in the dependency tree ($d_{S_i}$) and its parent or head ($h_{S_i}$). If the $d_{S_i}$ is more than a threshold[4], we check if there is an entity in the subtree whose root is $h_{S_i}$. If there is an entity we add it to the set $S_{E-depth}$ (e.g. *E1* in the subtree with root movie parent of oldest in Figure 2(B)). Otherwise, the superlative constraint should be applied as the last constraint. The type of each superlative constraint result will be the type of its immediate parent ($h_{S_i}$) in the tree.

*Comparative Constraint.* Comparative constraint is similar to the superlative constraint except that the depth threshold here is 3[5] (instead of 2) and the entity in the comparative keywords second parent (instead of immediate parent) subtree is extracted and added to $C_{depth}$. If there is no entity in the subtree then the comparative constraint is the last one to be applied (e.g. more in Figure 2(B)).

[4]Here 2 is selected by experiment.
[5]Selected based on experiments.

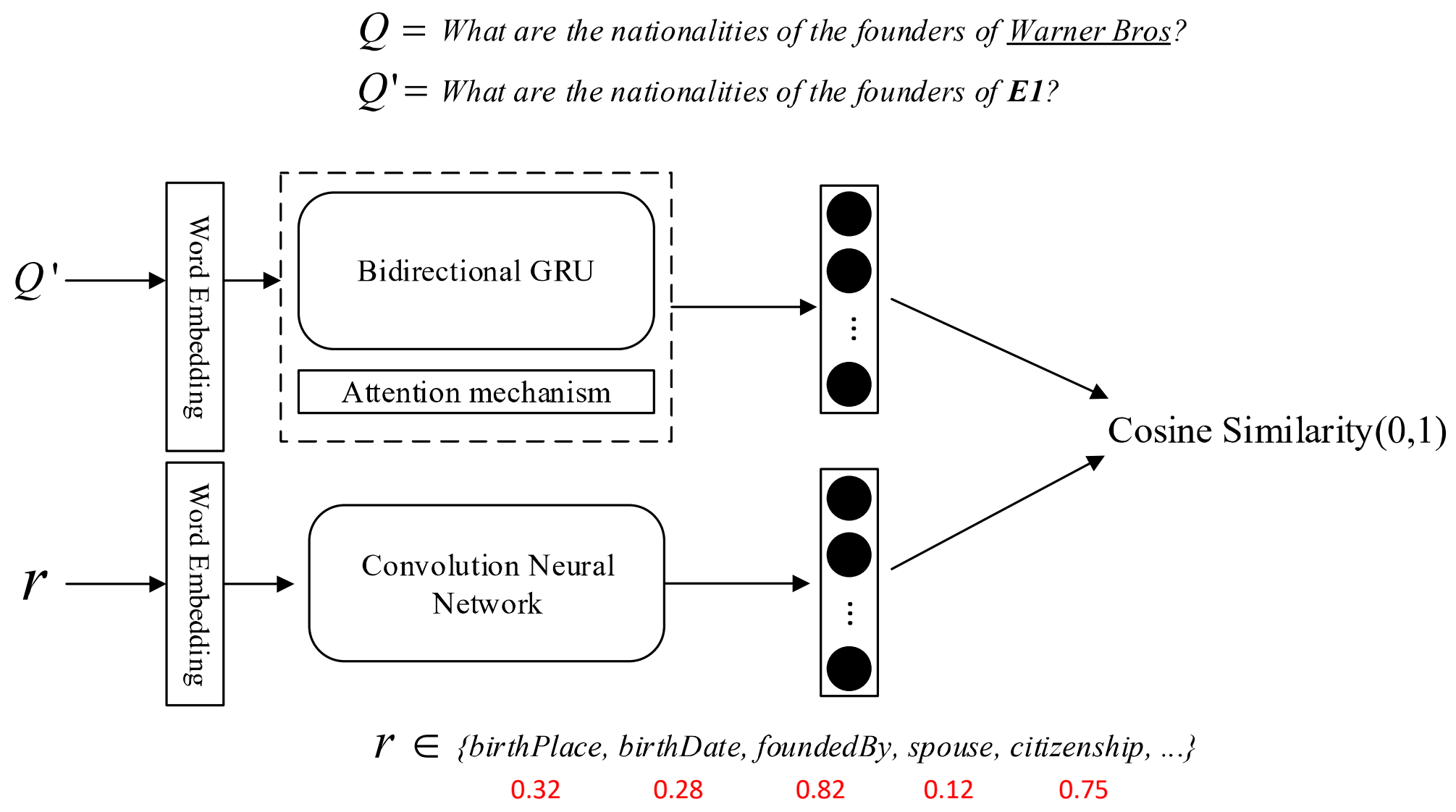

Figure 4: Semantic matching model for scoring relations based on input question.

*Multi-entity Constraint.* We split $E$, list of all entities found by NER, into two sets: $E_{depth}$ and $E_{topic}$. $E_{depth}$ contains the entities that are found in superlative, comparative, and implicit temporal (union of $C_{depth}$, $S_{E-depth}$, and implicit temporal) as we talked about. $E_{topic}$ holds the candidates of *topic entity* that are not in the depth ($E - E_{depth}$). In this paper, the topic entity is assumed as the main subject of $Q$ and the one which we need to go more than one hop from it to find the answer(s). Any entity in $E_{topic}$ can be chosen as the topic entity with the same probability. Each entity in $E_{topic}$ that is not selected as the topic entity is assumed as the entity constraint.

After detecting these constraints, we use Farsbase to extract triples for each constraint.

### 4.2. Logical Form Generation

In this part, we extract all of the triples up to two hops in these steps we create logical forms. To reduce our search space, we use a semantic matching model to prevent expanding relations with scores lower than a threshold. The inputs of semantic model are the word embedding matrix $E_{Q'} \in R^{|N|*d}$ of word sequence of $Q'$ (for all paraphrases in PeCoQ) and the word embedding matrix $E_r \in R^{|M|*d}$ of the candidate relation using fastText (Bojanowski et al., 2017). The model uses biGRU for $E_{Q'}$ with a self-attention mechanism and CNN for $E_r$. The output is the cosine similarity between feature vectors produced by GRU and CNN. Our model is similar to (Shin and Lee, 2020), however, we use CNN instead of using only a pooling layer on the relations present in the given question. The architecture of our model is shown in Figure 4.

After detecting constraints, it is time to create logical forms. Logical forms contain variables which are the unknown entities and represent both the answers and the hidden information required in the question. Since some constraints should be processed in sequence, we define priority for assembling. There is a set of variables $V$ that have been created during combination. The combination steps are as follows:

*Step1.*. Firstly, the relevant operational constraints are applied on entities in $C_{depth}$ and $S_{E-depth}$ are applied first. We consider operational phrases located in the depth of the dependency tree as a question with only one operational constraint. The answer is assumed as the topic entity and is added to $E_{topic}$ (e.g. *?v1* in Figure 3). To expand the variable later we use the set of relations of the same type of the variable in KG.

*Step2.*. We expand the selected topic entity from $E_{topic}$ using its corresponding triples (e.g. *(g2)* in Figure 3). The last variable of the hop and the middle variable for 2-hops are added to $V$. The last variable is selected as the answer node unless a comparative constraint is needed to be applied to it.

*Step3.*. Other constraints such as multi-entity constraint and explicit time are added to SPARQL by attaching them to the variables in $V$. The implicit temporal constraint is attached with a new relation to the variables connect to the topic entity.

*Step4.*. In this step the remaining operation constraints are applied to the answer. If there is a compare operation, the answer variable would be a new variable which is compared to the last relation in the topic entity's last relation (e.g. *step4* in Figure 3). The superlative operation would be applied to the variables in the relations of the topic entity and the aggregation would be applied only to the answer variable.

After finishing Step1 to Step4, a set of different SPARQLs is created. There is only one correct logical form in the set that its structure, number of hops, and the attachment variables (syntax), and its within relations (semantic) are correct. The SPARQLs that do not have answers are removed from the list. It is worth mentioning that entities in PeCoQ questions are not noisy. Therefore, from the entities that produced SPARQL the entity with the highest entity linking score (output score is between 0 and 1) is chosen.

### *4.3. Ranking and Selection*

To predict the best logical form/SPARQL from the candidates, we use Multilingual-BERT since the question is in Persian, and the relations in Farsbase and SPARQL are in English. The main challenge here is to choose the best logical form that best represents the question syntactically and semantically. Incorrect logical forms may have wrong relations or correct relations with wrong structure (according to the number of hops or variable attachments). The NN-based semantic matching model (Luo et al., 2018) performed poorly on PeCoQ. Therefore, we choose BERT as it encodes a rich hierarchy of linguistic information, syntactic features in the middle layers, and semantic features in the top layers (Jawahar et al., 2019). As Figure 5 shows, The inputs of Multilingual-BERT are the question $Q'$ and the candidate SPARQL that its entities are replaced with the same dummy tokens in the $Q'$ and the output is the similarity score in the range of [0,1].

For training, we use the noise-contrastive estimation (NCE) method. For each question, we generate the candidate set by applying our approach. The SPARQL in the set that has the most $F_1$-score among others is selected as the positive sample and other SPARQLs are the negative samples.

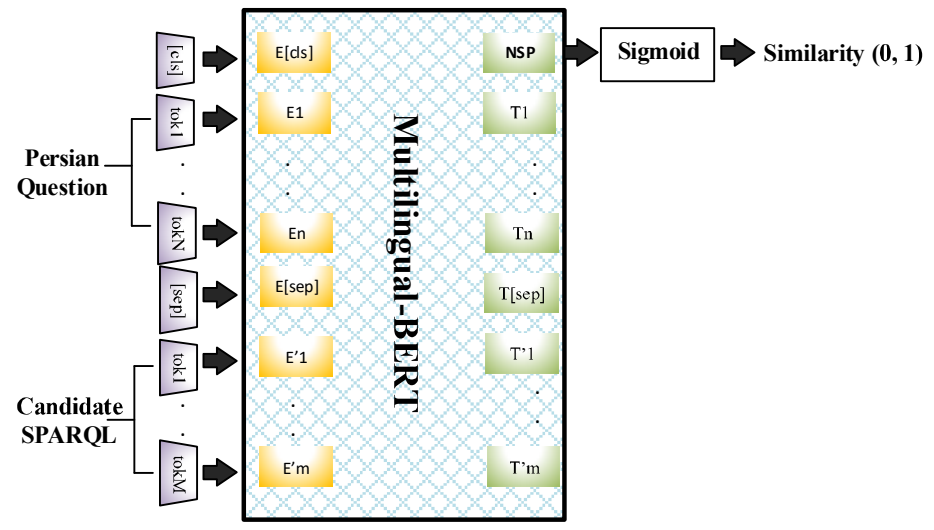

Figure 5: BERT model for ranking candidate SPARQLs

## 5. Experiments

We evaluate our KBQA approach on test data of PeCoQ with the average accuracy and F1-score as the evaluation metrics. Our method gains the accuracy of 62.75% and the F1-score of 62.98% in converting Persian questions into their corresponding logical forms. We also evaluate three main parts: NER tool, logical form generation, and logical form selection separately. Table 3 shows that fine-tuning ParsBert on PeCoQ outperforms the results of ParsBert trained on Arman, in the NER task. Since there are no previous works on PeCoQ, we compare our method to an end-to-end model that creates SPARQL of the given complex question. The hyperparameters of the model are changed so that it gives the best result on our Persian Dataset. Table all-res shows the results of evaluating our approach in creating logical forms. As the table shows, our method performs 8% better than the work by Zhang et al. (2019) in query generation on PeCoQ. In generating logical forms our method gains 70.72% accuracy and in selecting the correct form the accuracy of Multilingual-BERT is 88.73%. It means that in 70.72% of cases the correct logical form is generated among candidates, and totally in 62.75% of cases, the final selected logical form is correct. The total precision, recall, and F1 of our method are evaluated as 71.24%, 56.45%, and 62.98%, respectively. The results are shown in Table 2.

### *5.1. Error Analysis*

We randomly select 100 error cases and trace our method for producing corresponding logical forms manually.

*Entity Error:*. 7% errors are coming from our NER tool and 28% errors coming from not finding the correct entity from the knowledge graph with our entity linking model.

| Method | $F_1$-score | Precision | Recall | Accuracy |
|---|---|---|---|---|
| Zhang et al. (2019) | - | - | - | 54% |
| Ours/generation | - | - | - | 70.72% |
| Ours/total | 62.98% | 71.24% | 56.45% | **62.75%** |

Table 2: Results on PeCoQ. Ours/generation indicates the accuracy of logical forms generation only. Ours/total indicate the accuracy of our method selecting the best logical form as the answer

| NER Tool | Accuracy |
|---|---|
| ParsBERT-ARMAN | 51.74% |
| Fine-Tuned ParsBERT | **96.11**% |

Table 3: Results of NER on PeCoQ test set.

*Logical Form Generation Error:*. 24% of the time the correct logical form is not created due to dependency parser errors. Eliminating the correct relations with the semantic matching model leads to 10% of errors.

*Logical Form Selection:*. We observe that 12% of errors are related to the multi-lingual BERT and also 19% of the times the selected logical form cannot be queried against KG due to being uncanonical on some entities.

## 6. Conclusion

To the best of our knowledge, this is the first work on the Persian complex KBQA. We introduced a dependency parsing-based method to find the various constraints in the question, especially for those questions that have more than one operational constraint. Our method uses a semantic matching model to reduce the search space by eliminating relations with a score lower than a threshold. Possible candidate logical forms are created by rules based on the dependency tree, and then a Multilingual-BERT is used to select the logical form that best describes the input complex question syntactically and semantically. Experiments showed that BERT can capture semantical and syntactical clues to choose the best logical form among the set that have logical forms with different relations, hops, and structure.

Through our discussion, we considered multi-hop and multi-constraint questions. However, we restrict extracting information from Farsbase up to 2 hops. We believe that apart from using the semantic matching model for reducing the search space, future research should look for making the semantic matching model more powerful such that it can find the number of hops in the given question as well. In addition, future research should focus on developing PeCoQ dataset by adding more complex question.